\documentclass[conference]{IEEEtran}
\IEEEoverridecommandlockouts

\usepackage{cite}
\usepackage{amsmath,amssymb,amsfonts}
\usepackage{algorithmic}
\usepackage{graphicx}
\usepackage{textcomp}
\usepackage{xcolor}
\usepackage{booktabs} 
\usepackage{makecell} 
\def\BibTeX{{\rm B\kern-.05em{\sc i\kern-.025em b}\kern-.08em
    T\kern-.1667em\lower.7ex\hbox{E}\kern-.125emX}}
\begin{document}

\title{UMOD: A Novel and Effective Urban Metro Origin-Destination Flow Prediction Method}

\author{\IEEEauthorblockN{1\textsuperscript{st} Peng Xie}
\IEEEauthorblockA{\textit{School of Computing and Artificial Intelligence} \\
\textit{Southwest Jiaotong University}\\
Chengdu, China \\
pengxie@my.swjtu.edu.cn}
\and
\IEEEauthorblockN{2\textsuperscript{nd} Minbo Ma}
\IEEEauthorblockA{\textit{School of Computing and Artificial Intelligence} \\
	\textit{Southwest Jiaotong University}\\
	Chengdu, China \\
	minboma@my.swjtu.edu.cn}
\and
\IEEEauthorblockN{3\textsuperscript{rd} Bin Wang}
\IEEEauthorblockA{\textit{School of Computer Science and Technology} \\
\textit{Ocean University of China}\\
Qingdao, China \\
wangbin9545@ouc.edu.cn}
\and
\IEEEauthorblockN{4\textsuperscript{th} Junbo Zhang}
\IEEEauthorblockA{\textit{JD iCity, JD Technology} \\
\textit{JD Intelligent Cities Research}\\
Beijing, China \\
msjunbozhang@outlook.com}
\and
\IEEEauthorblockN{5\textsuperscript{th} Tianrui Li\textsuperscript{*}}\thanks{\textsuperscript{*}Corresponding Author.}
\IEEEauthorblockA{\textit{School of Computer Science and Technology} \\
\textit{Southwest Jiaotong University}\\
Chengdu, China \\
trli@swjtu.edu.cn}
}

\maketitle

\begin{abstract}
Accurate prediction of metro Origin-Destination (OD) flow is essential for the development of intelligent transportation systems and effective urban traffic management. Existing approaches typically either predict passenger outflow of departure stations or inflow of destination stations. However, we argue that travelers generally have clearly defined departure and arrival stations, making these OD pairs inherently interconnected. Consequently, considering OD pairs as a unified entity more accurately reflects actual metro travel patterns and allows for analyzing potential spatio-temporal correlations between different OD pairs. To address these challenges, we propose a novel and effective urban metro OD flow prediction method (UMOD), comprising three core modules: a data embedding module, a temporal relation module, and a spatial relation module. The data embedding module projects raw OD pair inputs into hidden space representations, which are subsequently processed by the temporal and spatial relation modules to capture both inter-pair and intra-pair spatio-temporal dependencies. Experimental results on two real-world urban metro OD flow datasets demonstrate that adopting the OD pairs perspective is critical for accurate metro OD flow prediction. Our method outperforms existing approaches, delivering superior predictive performance.
\end{abstract}

\begin{IEEEkeywords}
Origin-Destination Prediction, Urban Metro System, Spatio-Temporal AI, Spatio-Temporal Correlation, Traffic forecasting.
\end{IEEEkeywords}

\section{Introduction}
Accurate prediction of urban metro OD flow is vital for urban traffic management and intelligent transportation systems. This task involves developing a predictive model based on historical OD flow data to forecast future passenger flows between stations. By anticipating these flows, travelers can better plan their routes, while traffic management agencies and shared mobility platforms can optimize resource allocation and vehicle scheduling. Ultimately, such predictive capabilities contribute significantly to the development of a sustainable transportation system.

%


\begin{figure}[t]
	\centering
	\includegraphics[width=0.9\columnwidth]{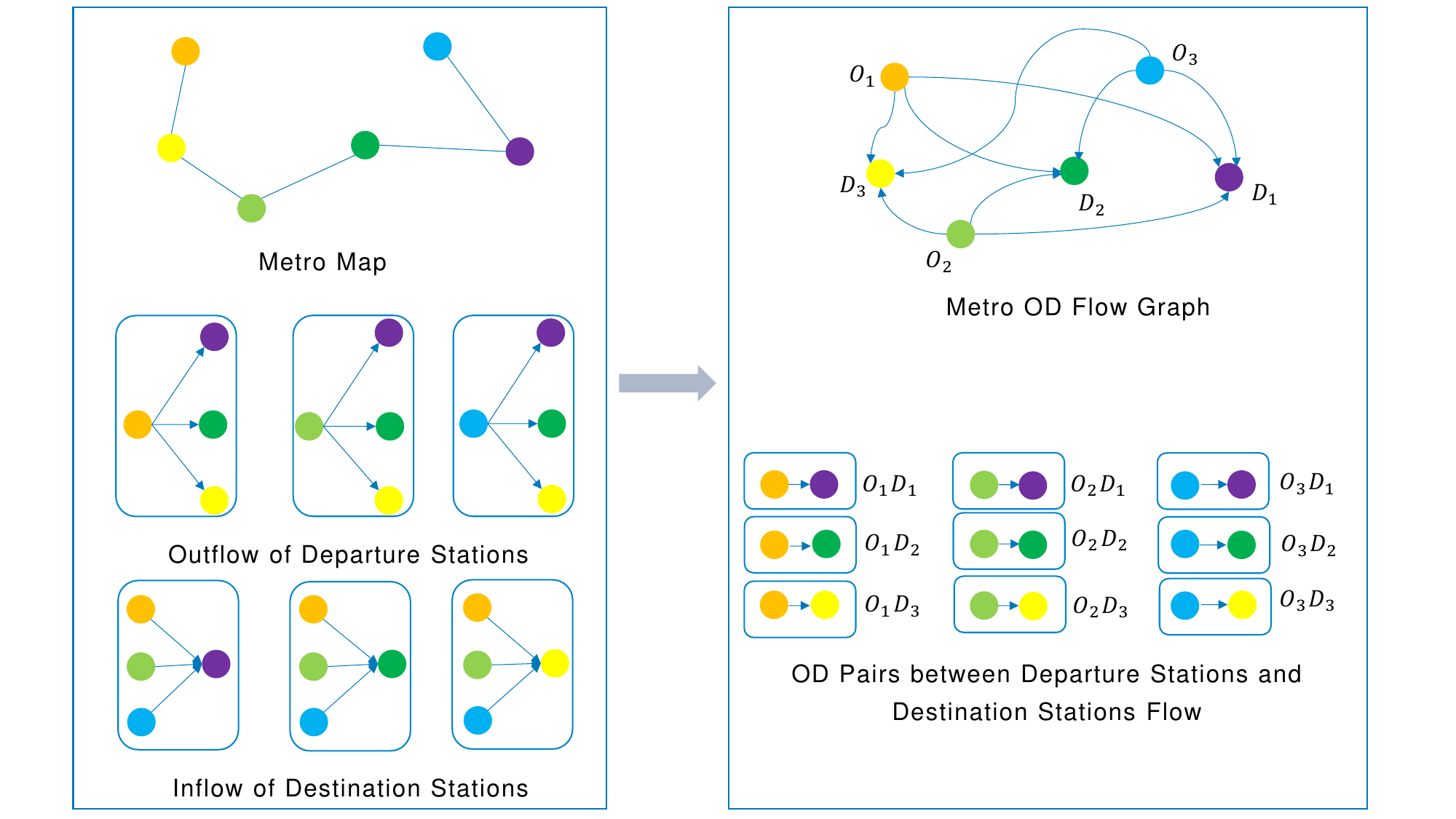} 
	\caption{The figure illustrates metro stations as interconnected nodes with two types of flows: Outflow and Inflow. By shifting to an origin-destination (OD) perspective, the OD flow graph in the upper right provides a comprehensive view of passenger movements, overcoming the limitations of traditional metro maps.}
	\label{OD_flow_v4}
\end{figure}

In the left portion of Fig. ~\ref{OD_flow_v4}, metro stations are depicted as interconnected nodes, allowing travelers to enter from any station and exit at another. Consequently, two distinct types of flows are illustrated on the lower side of the figure: Outflow and Inflow. Outflow, viewed from the departure station's perspective, highlights the stations to which passengers are headed. Conversely, Inflow, from the perspective of the Destination station, emphasizes the stations from which incoming passenger flows originate. By shifting our focus from a station-based departure/destination perspective to an origin-destination (OD) pairs perspective, we transcend the limitations imposed by the traditional metro map. This shift provides a bird's-eye view of the passenger flow patterns, culminating in the OD flow graph depicted in the upper right corner of the figure. Correspondingly, the lower left corner of the figure presents the OD pairs between Departure and Destination stations, offering further insights into the flow dynamics.

Extensive research has been conducted on metro OD flow prediction \cite{gong2020online, liu2023online, xu2023adaptive, zhu2023two}. Prior studies \cite{gong2020online, liu2023online, xie2023spatio} have predominantly focused on Outflow from the perspective of departure stations examining where the outgoing passenger flows are directed or Inflow from the perspective of destination stations analyzing the origin of passenger flows entering these stations. The spatial dependencies and flow distributions between stations have been studied through these two perspectives, as illustrated in the lower-left corner of Fig. ~\ref{OD_flow_v4}. Although these approaches have yielded significant insights and prediction accuracy for urban metro inflow and outflow, they overlook the inherent connection between departure and destination stations and the potential spatio-temporal relations between OD pairs.

To address these limitations, we propose a novel research perspective, referred to as the OD Pairs Flow View, as depicted in the lower right corner of Fig. ~\ref{OD_flow_v4}. This approach treats each trip's departure and destination stations as an inseparable unit. The key advantage of this perspective lies in its ability to accurately capture the dynamic flow changes between actual stations, allowing for a more detailed analysis of potential spatio-temporal relations between different OD pairs. Furthermore, this method is not constrained by specific metro network structures and applies to a variety of urban metro OD flow prediction and analysis scenarios.

Building on the motivations outlined above, we propose a novel and effective urban metro OD flow prediction method, UMOD. This method comprises three core modules: a data embedding module, a temporal relation module, and a spatial relation module. The data embedding module first transforms the raw OD pair input data into a latent space representation. This embedded representation is then fed into the temporal relation module, which captures the intra-pair temporal dependencies. Subsequently, it is processed by the spatial relation module to learn the inter-pair spatial dependencies. Ultimately, this sequence of operations yields the metro OD flow prediction results.

Our contributions can be summarized as follows:

\begin{itemize}
	\item We introduce UMOD, a groundbreaking urban metro OD flow prediction method that, for the first time, models travel demand from an origin-destination (OD) pairs perspective, offering a more accurate representation of actual passenger behavior. This novel approach shifts the focus from traditional station-based departure and destination views to a comprehensive bird's eye perspective, capturing the intricate dynamics of metro flow patterns in ways previously unexplored.
	
	\item To effectively capture the spatio-temporal relations of metro OD flow, we design three key modules: a data embedding module, a temporal relation module, and a spatial relation module.
	
	\item We conduct a comprehensive analysis to explore the impact of various historical time steps and prediction horizons on OD flow prediction performance. Our experiments reveal that, contrary to intuition, extending the historical time window does not necessarily lead to better prediction accuracy when the prediction horizon is fixed.
	
	\item Extensive experiments on two large-scale urban metro datasets demonstrate the significant predictive advantages of our proposed method. Compared to state-of-the-art approaches, UMOD achieves improvements in MAPE of 3.55\% and 18.34\% on SZ\_OD and CQ\_OD datasets, respectively, highlighting its superior performance.
\end{itemize}

The remainder of the paper is organized as follows. In Section 2, we introduce related work about traffic flow prediction and origin-destination prediction. In Section 3, we give symbol definitions and formalize the metro OD flow prediction problem. In Section 4, we show the overall framework of the proposed UMOD method. The experiment result, visualization, and analysis are given in Section 5. We conclude the work and show the future work in Section 6.

\section{Related work}
\subsection{Traffic Flow Prediction}\label{AA}
Accurate traffic flow prediction is of paramount importance for urban traffic management, urban planning, and public safety. In recent years, traffic prediction has garnered significant attention from both industry and academia, resulting in numerous advancements that have also impacted related fields such as economic management, earth science, and environmental science. Reflecting on the history of traffic flow prediction research, the methodologies can be broadly categorized into statistical learning-based, machine learning-based, and deep learning-based approaches.

Statistical learning methods include techniques like ARIMA (Autoregressive Integrated Moving Average) \cite{williams1998urban} and SARIMA (Seasonal Autoregressive Integrated Moving Average) \cite{zhang2011seasonal}. Subsequently, machine learning methods such as SVR (Support Vector Regression) \cite{lippi2013short} and K-NN (K-nearest neighbor) \cite{habtemichael2016short} emerged. However, these approaches often fail to account for the dynamic temporal and spatial dependencies inherent in traffic data. Recent studies \cite{zhang2017deep, li2018diffusion} have emphasized that capturing these temporal and spatial relations is crucial for enhancing the performance of traffic flow prediction.

With the rapid development of deep learning, there has been a growing adoption of deep learning-based methods for traffic flow prediction, and several reviews have documented the evolution of this field \cite{xie2020urban, tedjopurnomo2020survey, bui2022spatial}. Specifically, many deep learning approaches model traffic flow from the temporal dimension, the spatial dimension, or both simultaneously. In the temporal dimension, models such as LSTM \cite{hochreiter1997long}, GRU \cite{zhao2019t}, and Attention mechanisms \cite{guo2019attention, xu2020spatial, zheng2020gman} are commonly employed to capture temporal dependencies. In the spatial dimension, CNNs \cite{zhang2017deep} and GNNs \cite{li2018diffusion, yu2018spatio} are widely used to model spatial dependencies. Some studies have also focused on joint spatio-temporal modeling, with representative works including \cite{song2020spatial, wang2021gallat, wang2022hierarchical, xie2023spatio}. Additionally, other works explore capturing implicit spatial relations in traffic data \cite{bai2020adaptive, fang2021cdgnet, shao2022decoupled} and learning long-term temporal dependencies \cite{zheng2020gman, zhang2021transformer, ou2023stp}.

However, the existing research primarily captures spatial dependencies between urban areas, road nodes, and metro stations, as well as dynamic temporal dependencies, while overlooking the Origin-Destination (OD) patterns underlying traffic flow.

\subsection{Origin-Destination Prediction}
Origin-Destination prediction aims to forecast the traffic flow between specific areas or stations within a given time period. This task holds significant value for urban traffic managers in analyzing traffic flow distribution across the city and for travelers in planning their journeys in advance. As such, it has increasingly attracted the attention of researchers. \cite{rong2024interdisciplinary} provides a comprehensive review and analysis of OD flow modeling. Several deep learning-based spatio-temporal prediction methods have been developed specifically for OD flow prediction. For instance, a Contextualized Spatio-Temporal Network (CSTN) is designed for taxi OD flow prediction \cite{liu2019contextualized}. In \cite{wang2019origin}, the authors frame traffic demand forecasting as an Origin-Destination Matrix Prediction (ODMP) problem and propose a Grid-Embedding based Multi-task Learning (GEML) model. To capture the complex spatio-temporal dependencies for OD flow prediction, a Multi-Perspective Graph Convolutional Network (MPGCN) is introduced \cite{shi2020predicting}. Additionally, \cite{ke2021predicting} develope a spatio-temporal multi-graph convolutional network for predicting origin-destination ride-sourcing demand. \cite{han2022continuous} propose a continuous-time dynamic graph method for OD flow prediction, and more recently, \cite{xu2023adaptive} introduce a spatio-temporal adaptive feature fusion model tailored for metro OD flow prediction.

Despite these advancements, there remains a need for a simple yet effective urban metro flow prediction method that demonstrates robust predictive performance and adaptability across various urban metro networks.

\section{Problem Formulation}
Before introducing our method in detail, the symbol definitions and urban metro OD flow prediction task description are given. Table ~\ref{tab:notations} presents a list of commonly used notations and descriptions for reference. At time period $t$, the metro OD flow between two stations $i$ and $j$ can be expressed as $F_{t,i,j}$. The OD flow of the entire metro network can be expressed as $F_{t,:}\in \mathbb{R}^{N \times N}$, where $N$ means the number of metro stations. The metro OD flow prediction task can be defined as, given the history OD pair flow sequence, predicting the OD pair flow sequence for a period of time in the future.

\begin{equation}
	F_{t+1,:}, F_{t+2,:}, \dots, F_{t+P,:} = \mathcal{F}_{\theta} \left( F_{t,:}, F_{t-1,:}, \dots, F_{t-H+1,:} \right),
\end{equation}
where $\theta$ means all the learnable parameters in the UMOD model, $H$ is the length of the input OD flow sequence, and $P$ means the length of the predicted OD flow sequence.

\begin{table}[]
	\caption{Notations and descriptions.}
	\label{tab:notations}
	\centering
	\begin{tabular}{c|l}
		\hline
		\toprule
		\textbf{Notations} & \textbf{Description} \\
		\midrule
		$t$ & the current time interval \\ 
		$T$ & the length of the input OD flow sequence \\
		$P$ & the length of the predicted OD flow sequence \\ 
		$N$ & the number of metro stations \\ 
		$F_{t,:}\in \mathbb{R}^{N \times N}$  & the OD flow of the entire metro network \\ 
		$E_i \in \mathbb{R}^{T \times N \times d_i}$ & the input feature embedding presentation \\
		$E_a \in \mathbb{R}^{T \times N \times d_a}$ & the potential spatio-temporal dependency representation \\
		$E_h \in \mathbb{R}^{T \times N \times d_h}$ & the historical data embedding presentation \\
		$O_t \in \mathbb{R}^{T \times N \times d_h}$ & the output result of the temporal transformer module \\
		\bottomrule
	\end{tabular}
\end{table}

\section{Methodology}

\begin{figure}[t]
	\centering
	\includegraphics[width=0.9\columnwidth]{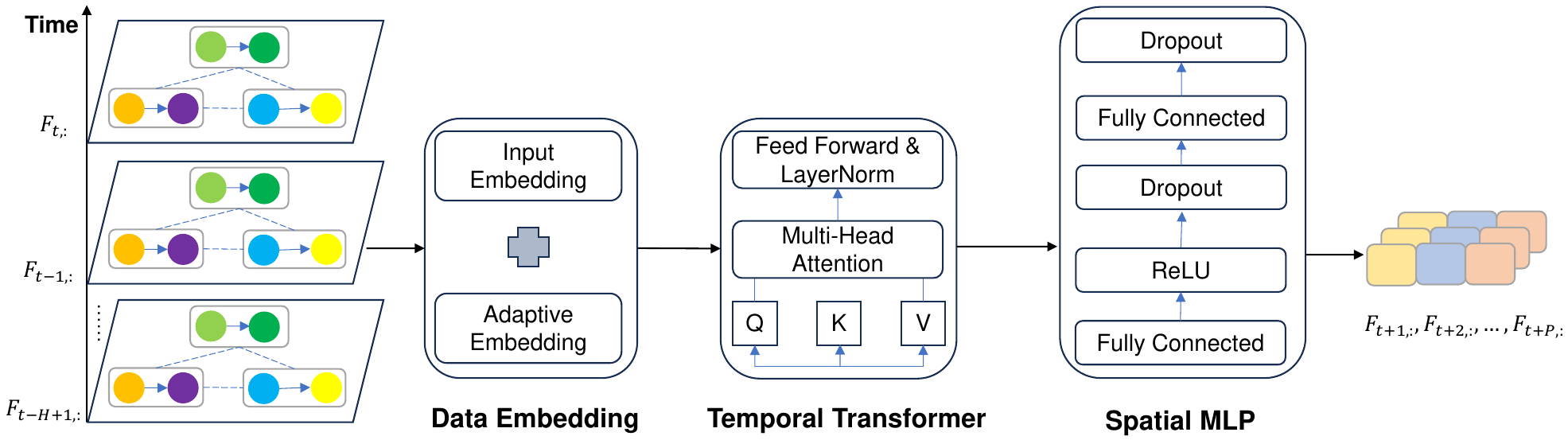} 
	\caption{The UMOD method consists of three core modules: a Data Embedding module, a Temporal Transformer module, and a Spatial MLP module.}
	\label{fig:UMOD_main_v4}
\end{figure}

The overall architecture of the model (UMOD) is shown in Fig. ~\ref{fig:UMOD_main_v4}. It contains a data Embedding module, a time series Transformer module, and a spatial MLP module.

\subsection{Data Embedding Module}
To project the raw OD pair inputs into hidden space representations, a data embedding module is designed. It includes Input Embedding and Adaptive Embedding. First, we use a fully connected layer to embed the input features of the historical OD flow data. The formula is as follows:

\begin{equation}
	E_i = FC(F_{t-H+1:t, :}),
\end{equation}

where $E_i \in \mathbb{R}^{T \times N \times d_i}$, $d_i$ means Input feature embedding dimensions, $FC(\cdot)$ is a fully connected layer.

Secondly, inspired by STAEformer \cite{liu2023spatio}, we use a spatio-temporal adaptive embedding layer to learn the potential spatio-temporal dependency representation $E_a \in \mathbb{R}^{T \times N \times d_a}$, where $d_a$ is the dimension of spatio-temporal adaptive embedding, $E_a$ is shared between different OD pairs.

Finally, we concatenate the input feature embedding $E_i$ with the spatio-temporal adaptive embedding $E_a$ to obtain the historical OD traffic data embedding $E_h \in \mathbb{R}^{T \times N \times d_h}$.

\subsection{Temporal Relation Module}

The Transformer model is used to capture intra-pair temporal dependencies in OD pairs sequence. First, the historical data embedding $E_h \in \mathbb{R}^{T \times N \times d_h}$ is input into the temporal Transformer layer to obtain the Query, Key, and Value matrices.

\begin{equation}
	\mathbf{Q} = \mathbf{E_h}\mathbf{W}_Q, \\
	\mathbf{K} = \mathbf{E_h}\mathbf{W}_K, \\
	\mathbf{V} = \mathbf{E_h}\mathbf{W}_V,
\end{equation}

where $\mathbf{W}_Q$, $\mathbf{W}_K$, and $\mathbf{W}_V$ are learnable parameters.

Then, the self-attention score is calculated as follows:

\begin{equation}
	A = \text{Softmax} \left( \frac{\mathbf{Q} \mathbf{K}^\top}{\sqrt{d_h}} \right),
\end{equation}

where $A \in \mathbb{R}^{N \times T \times T}$ captures the temporal dependency.

Finally, the output result of the temporal transformer module $O_t \in \mathbb{R}^{T \times N \times d_h}$ is obtained:

\begin{equation}
	O_t = \mathbf{A}\mathbf{V}.
\end{equation}

\subsection{Spatial Relation Module}

Unlike previous works that use CNN, GNN, or Transformer to capture spatial dependencies, we believe a two-layer MLP model can be used to capture the inter-pair spatial dependencies between OD pairs. The output of the temporal Transformer module is fed into the spatial MLP module to obtain the output result $\hat{Y} \in \mathbb{R}^{P \times N \times d_o}$, where $P$ is the predicted time steps and $d_o$ is the dimension of the output feature.

\section{Experiments}
In this section, we validate the effectiveness of our method using two real urban metro datasets. We begin by introducing the experimental datasets, environment configuration, and evaluation metrics. Following this, we compare our proposed UMOD method with both traditional models and the latest state-of-the-art (SOTA) approaches. We then perform ablation experiments to assess the contributions of the submodules within the UMOD method. Subsequently, we analyze the hyperparameters. Finally, we provide a discussion and further analysis of the experimental results.

\subsection{Experiment Settings}

1) Dataset description:

In this study, we use metro card swiping datasets from two cities in China to verify the effectiveness of our proposed UMOD method, namely the Shenzhen Metro dataset\cite{chen2022origin} and the Chongqing Metro dataset, as shown in Table~\ref{tab:datasets}.

Shenzhen Metro (SZ\_OD): The Shenzhen Metro card swiping data for 6 months from August 1, 2019 to January 31, 2020 is used, including 172 stations with a time granularity of 30 minutes.

Chongqing Metro (CQ\_OD): This dataset is private. It uses Chongqing Metro card-swiping data for 5 months from January 1, 2019 to May 31, 2019, covering 170 stations with a time granularity of 30 minutes.

\begin{table}[]
	\caption{The introduction of the two metro OD flow dataset.}
	\label{tab:datasets}
	\centering
	\begin{tabular}{cll}
		\toprule
		Dataset        & 	SZ\_OD    & 	CQ\_OD      \\
		\midrule
		City           & Shenzhen  & Chongqing        \\
		Time Span      & \makecell{2019.8.1 \\ -2020.1.31} & \makecell{2019.1.1 \\ -2019.5.31}  \\
		\# of Nodes    & 172 & 170   \\
		\bottomrule         
	\end{tabular}
\end{table}

2) Implementation details:
We use the PyTorch\cite{paszke2019pytorch} deep learning framework to implement the UMOD method proposed in the study, and use LibCity\cite{jiang2023libcity} to reproduce the comparison method, so as to achieve fair method comparison in a consistent environment. The GPU card is used in the experiment is NVIDIA GeForce RTX 4090, with a video memory of 24G. The Shenzhen Metro dataset is not processed additionally, and the dataset provided by the previous work\cite{chen2022origin} is directly used. The Chongqing Metro dataset removed the card-swiping data between 23:00-06:00 every day, because this time period is not during the operation period of the metro and no passengers enter or exit the stations. We use the data normalization method consistent with the AGCRN\cite{bai2020adaptive} method for the dataset. Both datasets are divided into training set, validation set, and test set in a ratio of 7:1:2. The batch size is 16. We use the Adam\cite{kingma2014adam} optimizer to optimize our model, set the maximum number of epochs to 100, and adopt an early stopping strategy, a patience value of 20, and a learning rate of 0.001. We use the historical OD traffic data of the past hour to predict the OD traffic data for the next hour.

3) Evaluation metrics:
We used three evaluation indicators commonly used in spatio-temporal prediction tasks to evaluate our method, i.e. Mean Absolute Error (MAE), Root Mean Square Error (RMSE), and Mean Absolute Percentage Error (MAPE). The formulas are as follows.

\begin{itemize}
	\item MAE
	
	\begin{equation}
		MAE=\frac{1}{n} \sum_{i=1}^{n}\left|\hat{y}_{i}-y_{i}\right|.
	\end{equation}
	
	\item RMSE
	
	\begin{equation}
		RMSE=\sqrt{\frac{1}{n} \sum_{i=1}^{n}\left(\hat{y}_{i}-y_{i}\right)^{2}}.
	\end{equation}
	
	\item MAPE
	
	\begin{equation}
		MAPE = \frac{100\%}{n} \sum_{i=1}^{n} \left| \frac{\hat{y}_i - y_i}{y_i} \right|,
	\end{equation}
\end{itemize}
where $n$ is the number of test samples, $\hat{y}_i$ and $y_i$ are the predicted site OD flow and the actual site OD flow, respectively. $\hat{y}_i$ and $y_i$ are converted to the original value scale through reversible Z-score normalization and then calculated.

4) Baselines:

We compared the UMOD method with other seven methods, as shown in Table \ref{tab:main_results_sz} and Table \ref{tab:main_results_cq}. These models can be divided into four categories, including (1) classic deep learning methods: MLP; (2) classic urban OD flow prediction methods: GEML; (3) classic spatio-temporal prediction methods: STID; (4) the latest representative multivariate time series prediction methods in recent years: DLinear, TSMixer, TimeMixer, SOFTS. The detailed introduction of these models is as follows:

\begin{itemize}
	\item MLP\cite{rumelhart1986learning}: We use a multi-layer perceptron network (MLP), a classic feed-forward neural network, to capture the complex nonlinear relations of OD flow data.
	
	\item GEML\cite{wang2019origin}: This is a classic Grid-Embedding-based OD traffic matrix prediction method that considers the fusion of spatio-temporal features.
	
	\item STID\cite{shao2022spatial}: A classic MLP-based efficient and concise spatio-temporal prediction model based on the characteristics of spatio-temporal indistinguishability is designed, and its prediction performance exceeded that of most spatio-temporal graph neural networks at that time.
	
	\item DLinear\cite{zeng2023transformers}: It considers whether the Transformer-based model is suitable for long-term time series prediction problems, and proposes a simple multivariate time series linear prediction model.
	
	\item TSMixer\cite{chen2023tsmixer}: This work proposes a time series prediction method with a pure MLP structure.
	
	\item TimeMixer\cite{wang2024timemixer}: This is a time series prediction method based on the idea of multi-scale mixing, which has achieved good experimental results in both long-term and short-term predictions.
	
	\item SOFTS\cite{han2024softs}: This work designs a novel STar Aggregate-Redistribute (STAR) module to capture the dependencies between channels, and proposes an MLP-based Series-cOre Fused Time Series forecaster for multivariate time series forecasting.
	
\end{itemize}
	
\subsection{Experimental Results}
Table 2 shows the prediction results of our proposed method UMOD and seven comparison methods on the Shenzhen Metro dataset. The historical OD flow data of the past hour is used to predict the OD flow data for the next hour. The three evaluation indicators of MAE, RMSE, and MAPE are used. It can be seen that our method is ahead of the comparison methods in these three evaluation indicators. Compared with the second-best comparison method, the improvement of our proposed UMOD method in the three evaluation indicators of MAE, RMSE, and MAPE is 20.22\%, 20.63\%, and 3.55\%, respectively. Specifically, the GEML method performs the worst on the Shenzhen Metro dataset. Since the GEML method is a multi-task learning framework, it predicts the OD flow and inflow/outflow flow of the grid. It takes OD flow prediction as the main task and inflow/outflow flow prediction as two subtasks, and experiments are conducted on the online car-hailing dataset. However, we believe that the OD flow pattern of online car-hailing travel is significantly different from that of the subway, so it can be seen that the GEML method performs poorly. In addition, we also compare the classic spatio-temporal prediction STID method and find that its experimental results are better than those of MLP, DLinear, and TSMixer methods. Finally, we also compare the multivariate time series prediction methods SOFTS and TimeMixer, which have performed well in both long-term and short-term time series prediction in recent years. Our proposed method still maintains good prediction performance.

We can find that the GEML method also performs the worst in all evaluation indicators by observing the experimental results on the Chongqing metro dataset. In terms of the MAE evaluation indicator, STID achieves the best prediction performance results, which may be due to its ability to capture the spatio-temporal dependencies in the data. In terms of the RMSE evaluation indicator, DLinear achieves the best prediction performance results, indicating that the spatial dependencies between the OD pairs of Chongqing metro are relatively weak, and the channel-independent multivariate time series method can achieve good prediction results. In terms of the MAPE indicator, the method we proposed, UMOD, achieves the best prediction performance results, with an improvement of 18.34\% compared to the second-best method, STID. While our UMOD method does not achieve the best performance in terms of the MAE and RMSE evaluation metrics, it proves to be the most competitive when considering overall prediction results and improvement ratios. Specifically, the top-performing method in MAE and RMSE only outperforms the second-best method by marginal increments of 0.69\% and 0.02\%, respectively, indicating minimal gains. In contrast, our UMOD method significantly enhances predictive performance in terms of the MAPE metric, achieving an impressive 18.34\% improvement over the second-best method.

\begin{table}[ht]
	\caption{The result of Shenzhen Metro OD flow (input 2-output 2) prediction}
	\label{tab:main_results_sz}
	\centering
	\begin{tabular}{clllll}
		\hline
		\multicolumn{1}{l}{\textbf{Dataset}} & \textbf{Methods}         & \textbf{MAE}                        & \textbf{RMSE}                        & 
		\textbf{MAPE (\%)}
		\\
		\hline
		& MLP                     & 2.189
		
		& 6.838
		
		& 60.20
		
		\\
		& GEML                 & 4.240
		
		& 15.729
		
		& 121.30
		
		\\
		& STID                      & \underline{1.825}
		
		&  5.530
		
		& \underline{54.42}
		
		\\
		& DLinear                     & 2.503
		
		& 7.226
		
		& 76.40
		
		\\
		SZ\_OD & TSMixer                     & 2.492
		
		& 7.274
		
		& 75.54
		
		\\
		
		& TimeMixer                     & 1.913
		
		& \underline{4.460}
		
		& 72.90
		
		\\
		& SOFTS                     & 2.556
		
		& 7.490                               & 78.43
		\\
		& UMOD                      & \textbf{1.456}
		&  \textbf{3.540}
		
		& \textbf{52.49}
		\\ 
		& Improvement (\%)                      & 20.22
		&  20.63
		
		& 3.55
		\\
		\hline
	\end{tabular}
\end{table}

\begin{table}[ht]
	\caption{The result of Chongqing Metro OD flow (input 2-output 2) prediction}
	\label{tab:main_results_cq}
	\centering
	\begin{tabular}{clllll}
		\hline
		\multicolumn{1}{l}{\textbf{Dataset}} & \textbf{Methods}         & \textbf{MAE}                        & \textbf{RMSE}                        & 
		\textbf{MAPE (\%)}
		\\
		\hline
		& MLP                     & 2.211
		
		& 6.680
		
		& 68.01
		
		\\
		& GEML                 & 3.486
		
		& 7.267
		
		& 208.29
		
		\\
		& STID                      & \textbf{1.806}
		
		&  6.672
		
		& \underline{36.87}
		
		\\
		& DLinear                     & 1.821
		
		& \textbf{6.612}
		
		& 37.28
		
		\\
		CQ\_OD & TSMixer                     & 1.821
		
		& \underline{6.613}
		
		& 37.23
		
		\\
		
		& TimeMixer                     & 1.823
		
		& 6.615
		
		& 37.81
		
		\\
		& SOFTS                     & \underline{1.819}
		
		& 6.686                               & 37.18
		\\
		& UMOD                      & 1.917
		&  6.859
		
		& \textbf{30.11}
		\\ 
		& Improvement (\%)                      & 0.69
		&  0.02
		
		& 18.34
		\\
		\hline
	\end{tabular}
\end{table}

\begin{table}[ht]
	\caption{The prediction result of Shenzhen Metro OD flow with variable historical (H) and prediction (P) window size}
	\label{tab:variable_H_P}
	\centering
	\begin{tabular}{cllll}
		\hline
		\multicolumn{1}{l}{\textbf{Datasets}} & \textbf{H-P}         & \textbf{MAE}                        & \textbf{RMSE}                        & 
		\textbf{MAPE} \\
		\hline
		&  6-2                    & \textbf{1.429}                                 & \textbf{3.311}                               & \textbf{51.59\%}  \\
		& 6-4                 & 1.914                                 & 6.218                                & 81.63\%  \\
		& 2-2                      & \underline{1.456}                                &   \underline{3.540}                             & \underline{52.49\%}    \\
		SZ\_OD & 4-4                     & 1.789                                 & 5.222        & 72.38\%     \\
		& 6-6       &        1.740      & 4.825           & 67.91\%      \\
		& 2-4                     & 1.703          & 4.738                                & 64.08\%        \\
		& 2-6                      & 1.845                              &  5.651                                & 69.74\%         \\ 
		\hline
	\end{tabular}
\end{table}

Furthermore, to assess the impact of varying historical and future time steps on prediction accuracy, we conduct experiments by adjusting these parameters to observe their influence on the prediction results. We divide the historical time step (H) and the future time step (P) into three categories, the first category is H>P, the second category is H=P, and the third category is H<P. As shown in Table~\ref{tab:variable_H_P}, our model UMOD shows the best prediction effect both in the past three hours and the next one hour, which was shown in the experiment set that the historical time step is 6 and the future time step is 2, respectively. The worst prediction performance occurs when using a historical time step of 6 and a future time step of 4, meaning that the model predicts the next two hours of data based on the previous three hours of historical data. In addition, we also find that it is not always the case that the longer the historical time step, the better the effect when the future prediction time step is fixed. For instance, the model achieves better prediction accuracy with a historical time step of 2 compared to historical time steps of 4 or 6 when the prediction time step is set to 4. We believe that a longer historical time step may also introduce some noise and interfere with the prediction results. More broadly, determining the optimal historical time step and future prediction step based on the specific characteristics of the data remains an open research question that warrants further investigation.

\subsection{Ablation Study}
We design an ablation experiment to verify the impact of different embedding methods on the metro OD flow prediction results. First, we removed the Input Embedding module and named the ablation model UMOD w/o $E_i$. Second, we removed the Adaptive Embedding module and named this ablation model UMOD w/o $E_a$. Finally, we compared the previous two ablation models with the proposed model UMOD. The experimental results on the Shenzhen metro flow dataset are shown in Table~\ref{tab:albation_study}.

\begin{figure}[t]
	\centering
	\includegraphics[width=0.9\columnwidth]{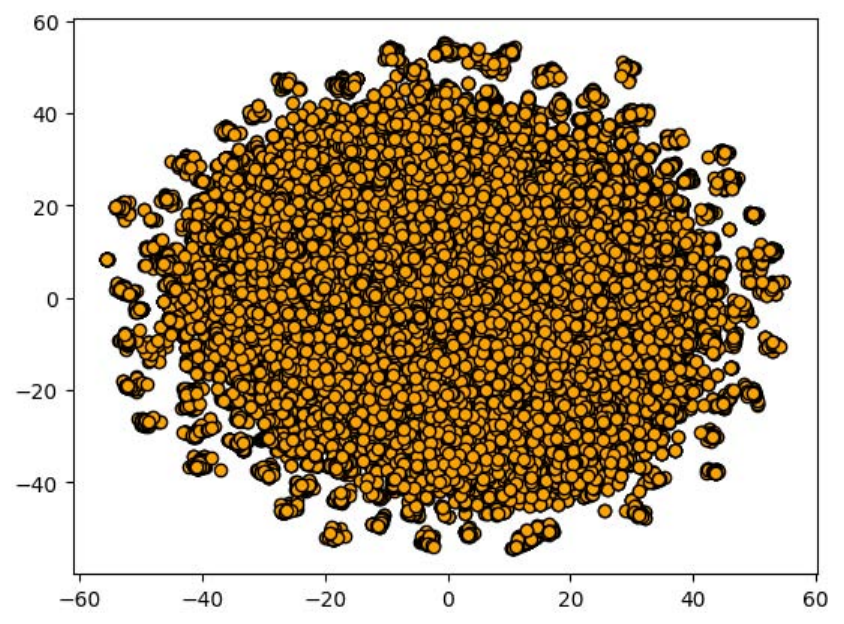} 
	\caption{Spatial embedding (OD pairs embedding) of OD flow.}
	\label{fig:Spatial_embedding}
\end{figure}

From the experimental results, we can see that removing the Input Embedding module has the largest decrease in the prediction results of the model. Secondly, removing the Adaptive Embedding module has a slight decrease in the prediction results of the model compared with the UMOD model. By integrating the Input Embedding and Adaptive Embedding modules, the prediction performance of the UMOD model is the best overall. We visualize the spatial feature embedding representation, as shown in Figure~\ref{fig:Spatial_embedding}. As can be seen from the figure, a large number of OD pairs are clustered together, while a small number of OD pairs are scattered in small groups. We believe that OD pairs reflect people's travel needs. Since people's travel is regular and group-oriented, there are a large number of clustered phenomena in spatial distribution, while OD pairs scattered around may be some with low travel frequency or affected by personal travel preferences, and there is a certain degree of uncertainty or randomness. This observed phenomenon is also more in line with the characteristics and rules of people's urban travel.

\begin{table}[ht]
	\caption{Analysis of albation study of Shenzhen Metro OD flow (input 2-output 2) prediction}
	\label{tab:albation_study}
	\centering
	\begin{tabular}{llll}
		\hline
		\textbf{Methods}         & \textbf{MAE}                        & \textbf{RMSE}                        & 
		\textbf{MAPE}
		\\
		\hline
		UMOD w/o E\_i                     & 2.193
		
		& 7.757
		
		& 62.64\%
		
		\\
		
		UMOD w/o E\_a                     & 1.555
		
		& 4.429
		
		& \textbf{49.96\%}
		
		\\
		UMOD                      & \textbf{1.456}
		&  \textbf{3.540}
		
		& 52.49\%
		\\ 
		\hline
	\end{tabular}
\end{table}

\subsection{Hyper-parameters Analysis}
\begin{figure}[t]
	\centering
	\includegraphics[width=0.9\columnwidth]{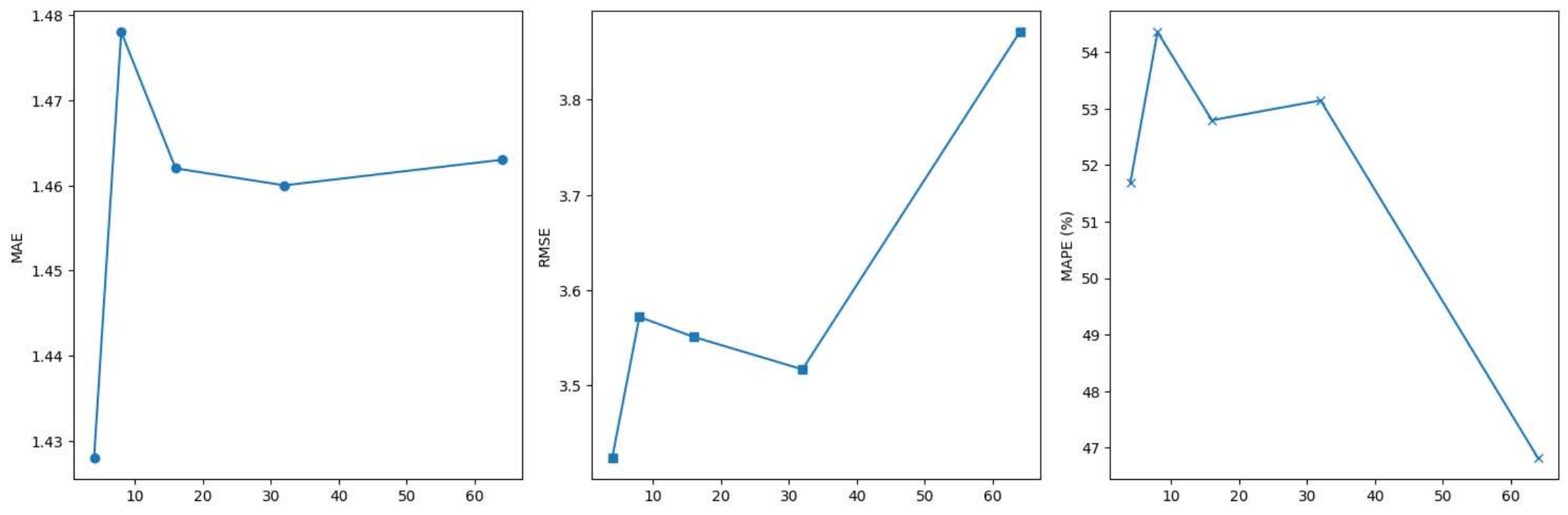} 
	\caption{Impact of Number of Input Embedding Dimensions.}
	\label{fig:y_i_embedding}
\end{figure}

\begin{figure}[t]
	\centering
	\includegraphics[width=0.9\columnwidth]{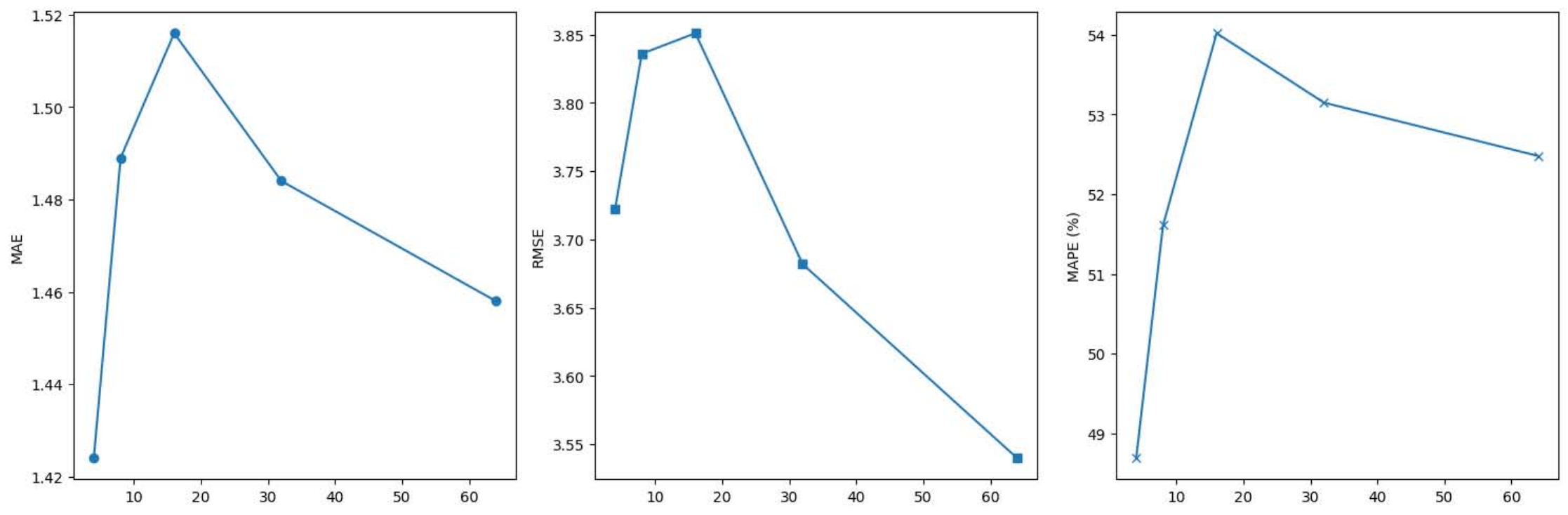} 
	\caption{Impact of Number of Adaptive Embedding Dimensions.}
	\label{fig:y_a_embedding}
\end{figure}

We analyse the impact of the Input Embedding and Adaptive Embedding dimension sizes on the model prediction results. The dimension size of Adaptive Embedding is fixed to 80, and the selected Input Embedding dimension sizes were 4, 8, 16, 32, and 64. The dimension size of Input Embedding is fixed to 24, and the selected Adaptive Embedding dimension sizes were 4, 8, 16, 32, and 64. The experimental results are shown in Figures ~\ref{fig:y_i_embedding} and ~\ref{fig:y_a_embedding}. Figure ~\ref{fig:y_i_embedding} shows that the MAE and RMSE values are the smallest when the Input Embedding dimension is 4. The MAPE value is the smallest when the Input Embedding dimension is 64. The size of the Input Embedding dimension has a certain impact on the model prediction results. It is not the case that the larger the dimension, the better the model prediction results. In Figure ~\ref{fig:y_a_embedding}, the MAE value and MAPE values are the smallest when the Adaptive Embedding dimension is 4. The RMSE value is the smallest when the Adaptive Embedding dimension is 64. Different Adaptive Embedding dimensions have different effects on the model prediction results. As the Adaptive Embedding dimension increases, the model prediction results first get worse and then get better. The prediction effect is the worst when the Adaptive Embedding dimension is 16. Determining the optimal dimensions for Input Embedding and Adaptive Embedding remains an area for further exploration and analysis in future research.

\subsection{Discussion}
In this study, we use metro datasets from Shenzhen and Chongqing in China to verify the effectiveness of the proposed UMOD method and compare it with seven models including classic deep learning models, spatio-temporal prediction models, and the latest multivariate time series prediction models. The experimental results show that our method has a significant advantage in improving prediction performance compared with the comparison methods. To verify the impact of different historical time steps and future time steps on the prediction results, we also conduct experiments to set different historical time steps and future time steps to see their impact on the prediction results. This approach has rarely been explored and analyzed in the field of traffic prediction and spatio-temporal prediction before, which is of great value for discovering and understanding the impact of data characteristics on prediction results. In addition, we also set up ablation experiments to analyze the impact of different data embedding modules on the metro OD flow prediction results, and visualize and explain the spatial feature embedding representation, which helps people make better traffic travel decisions. Finally, to analyze the impact of hyperparameters on the model prediction results, we also set different dimension sizes for Input Embedding and Adaptive Embedding to analyze their impact on the prediction results.

\section{Conclusions and Future Work}
In this paper, we propose a simple yet effective urban metro OD flow prediction method, UMOD, which integrates a data embedding module, a temporal Transformer module, and a spatial MLP module. We conducted experiments on two metro datasets from Shenzhen and Chongqing, China. The experimental results demonstrate that our proposed UMOD method offers significant predictive performance advantages and effectively captures the complex dynamic spatiotemporal dependencies inherent in metro OD flow data. To validate the effectiveness of the data embedding module, we performed corresponding ablation experiments and hyperparameter analyses. Additionally, to enhance our understanding of the data characteristics and optimize prediction results for traffic decision-making, we investigated the impact of varying historical and future time steps on the performance of metro OD flow prediction.

In the future, we aim to further analyze the underlying patterns and principles of urban travel, employing data and knowledge fusion methods to enhance the credibility and interpretability of model predictions, thereby providing better support for traffic decision-making. Additionally, we will investigate how to select optimal historical and future prediction time steps based on data characteristics to maximize the model's effectiveness. Given the strong few-shot learning capabilities of large language models, we also plan to explore their application in traffic prediction and decision-making across different regions and cities.

\section*{Acknowledgment}
This research was supported by the National Natural Science Foundation of China (Nos. 62176221, 61572407), and Sichuan Science and Technology Program (Nos. 2024NSFTD0036, 2024ZHCG0166). The first author would like to personally express my sincere gratitude to Yanyu Yao from Zhejiang Wanli University for her invaluable assistance in polishing the language of this paper.

\bibliography{mybibfile}

\bibliographystyle{IEEEtran}

\end{document}